\documentclass[letterpaper, 10 pt, conference]{ieeeconf}  

\IEEEoverridecommandlockouts                              

\usepackage{graphics} 
\usepackage{epsfig} 
\usepackage{mathptmx} 
\usepackage{times} 
\usepackage{amsmath} 
\usepackage{amssymb}  
\usepackage{cite} 

\title{\LARGE \bf
Overcoming Visual Clutter in Vision Language Action Models via Concept-Gated Visual Distillation
}

\author{Sangmim Song$^{1}$, Sarath Kodagoda$^{1}$, Marc Carmichael$^{1}$ and Karthick Thiyagarajan$^{2}$
\thanks{$^{1}$Sangmim Song, Sarath Kodagoda, and Marc Carmichael are with the University of Technology Sydney, NSW, Australia
        {\tt\small Sangmim.song@uts.edu.au }}%
\thanks{$^{2}$Karthick Thiyagarajan is with Western Sydney University, NSW, Australia
        {\tt\small K.Thiyagarajan@westernsydney.edu.au}}%
}

\begin{document}

\maketitle
\thispagestyle{empty}
\pagestyle{empty}

\begin{abstract}
Vision-Language-Action (VLA) models demonstrate impressive zero-shot generalization but frequently suffer from a "Precision-Reasoning Gap" in cluttered environments. This failure is driven by background-induced feature dilution, where high-frequency semantic noise corrupts the geometric grounding required for precise manipulation. To bridge this gap, we propose Concept-Gated Visual Distillation (CGVD), a training-free, model-agnostic inference framework that stabilizes VLA policies. CGVD operates by parsing instructions into safe and distractor sets, utilizing a two-layer target refinement process—combining cross-validation and spatial disambiguation—to explicitly penalize false positives and isolate genuine manipulation targets. We then process the scene via Fourier-based inpainting, generating a clean observation that actively suppresses semantic distractors while preserving critical spatial geometry and visual proprioception. Extensive evaluations in highly cluttered manipulation tasks demonstrate that CGVD prevents performance collapse. In environments with dense semantic distractors, our method significantly outperforms state-of-the-art baselines, achieving a 77.5\% success rate compared to the baseline's 43.0\%. By enforcing strict attribute adherence, CGVD establishes inference-time visual distillation as a critical prerequisite for robust robotic manipulation in the clutter.
\end{abstract}

\section{INTRODUCTION}

The pursuit of general-purpose robotic manipulation has been fundamentally accelerated by the advent of Vision-Language-Action (VLA) models~\cite{openvla, rt2, octo, gr00t, pi0}. By grounding large language models into robotic control policies, these architectures have demonstrated remarkable zero-shot generalization capabilities, enabling robots to follow open-vocabulary instructions like `'Put spoon on towel'' without task-specific training. These models promise a future where robots can operate in unstructured, human-centric environments.

\begin{figure}[t]
    \centering
    \includegraphics[width=\columnwidth]{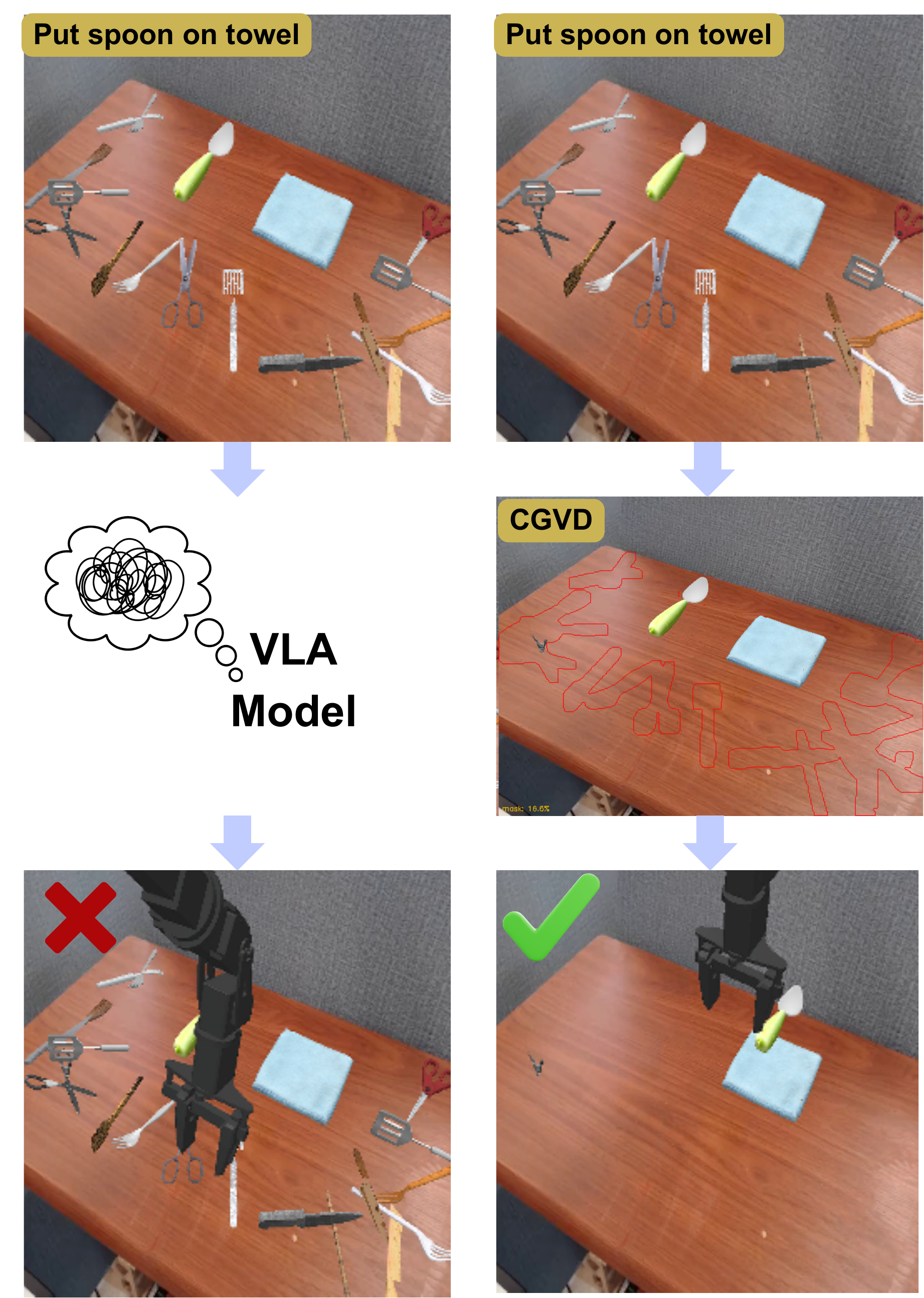}
    \caption{Comparison of manipulation task execution in cluttered environments. While a standard VLA model (left) struggles with object confusion in a highly cluttered scene, our CGVD approach (right) successfully identifies and places the target object ("spoon") on the towel. }
    \label{fig:full_column_image}
\end{figure}
However, a significant gap remains between the semantic reasoning capabilities of these models and their geometric precision in deployment conditions. While VLAs excel in curated, clutter-free environments, their performance degrades precipitously in the presence of visual clutter~\cite{distracted_robot, eva_vla}. We term this phenomenon the Precision-Reasoning Gap: the model successfully identifies the target object conceptually, yet attention corruption from surrounding distractors dilutes the latent representation used for spatial planning~\cite{obeyed}. This feature dilution manifests as high-variance trajectories, hesitation near distractors, and ultimately manipulation failure.

Critically, this degradation is not uniform across distractor types. We observe that failure concentrates around distractors sharing visual or semantic properties with the target~\cite{distracted_robot}. While VLAs may exhibit resilience to arbitrary clutter via large-scale pre-training, they remain brittle to semantically confusable objects; for example, a fork scattered near a target spoon triggers conflicting visual tokens within the same affordance category, causing the policy to attend to or even grasp the wrong object.

\begin{figure*}[t]
    \centering
    \includegraphics[width=\textwidth]{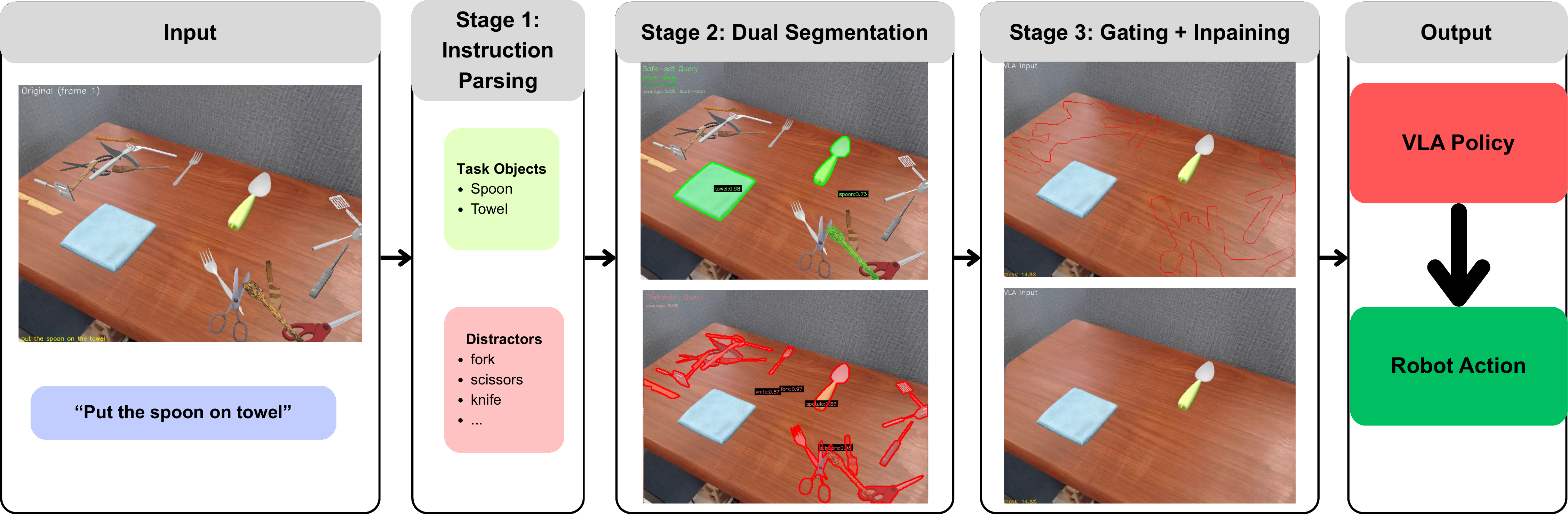}
    \caption{Overview of the CGVD pipeline. \textbf{Stage 1:} The language instruction is parsed to extract a safe set tar and a distractor set. \textbf{Stage 2:} SAM3 segments both sets independently, producing a safe-set mask and a distractor mask via dual-channel segmentation. \textbf{Stage 3:} Set-theoretic gating subtracts the safe-set mask from the distractor mask, and LaMa inpaints the resulting regions to produce a distilled observation passed to the VLA policy.}
    \label{fig:methodology}
\end{figure*}

Existing approaches to mitigate clutter-induced failure fall into three primary categories. Adaptation methods, such as OBEYED-VLA~\cite{obeyed}, fine-tune attention adapters to focus on targets. While effective, this requires expensive, architecture-specific retraining and limits generalization to the fine-tuning distribution. Inference-time intervention methods, such as BYOVLA~\cite{byovla}, use a VLM to identify distractors and a sensitivity probe to determine which to remove. However, this approach relies on external API calls (GPT-4o), requires multiple VLA forward passes per region for its probe, and provides only probabilistic protection the target can still be modified if both the VLM and the sensitivity threshold fail to flag it. Training-time augmentation methods~\cite{rosie, genaug, nice} generate diverse cluttered training data via generative models, improving robustness at the cost of retraining and without guarantees at deployment.

To bridge this gap without the cost of retraining or the fragility of existing inference-time approaches, we propose Concept-Gated Visual Distillation (CGVD): a model-agnostic inference framework that leverages modern vision foundation models~\cite{sam3} to selectively restructure visual observations before they reach the VLA policy. CGVD parses the task instruction to identify target and anchor objects, segments both distractors and task-relevant entities independently and uses set-theoretic subtraction to produce a distractor mask from which the target is architecturally excluded. Distractors are then replaced via inpainting~\cite{lama}, preserving the context of the scene.

Our contributions are as follows:
\begin{itemize}
    \item \textbf{Concept-Gated Visual Distillation (CGVD):} We introduce a training-free, model-agnostic inference framework that selectively removes distractors from VLA observations via language-grounded segmentation and inpainting, while preserving scene context.

    \item \textbf{Interaction-Aware Masking Logic:} To overcome the inherent semantic confusion of open-set vision foundation models which evaluate text prompts independently, we propose a set-theoretic cross-validation pipeline. This logic mathematically penalizes false positives and uses spatial disambiguation to isolate true targets from visually confusing distractors,

    \item \textbf{Demonstrated Clutter Robustness at Scale:}  We systematically evaluate CGVD on state-of-the-art VLAs ($\pi_0$, GR00T) within the SimplerEnv benchmark. Our framework prevents policy collapse in highly cluttered scenes, achieving up to a 77.5\% success rate compared to the baseline’s 43.0\% in semantic clutter, and demonstrates superior zero-shot adherence to complex attribute prompts.
\end{itemize}

\section{Related Work}
\begin{figure*}[t]
  \centering
  \includegraphics[width=\textwidth]{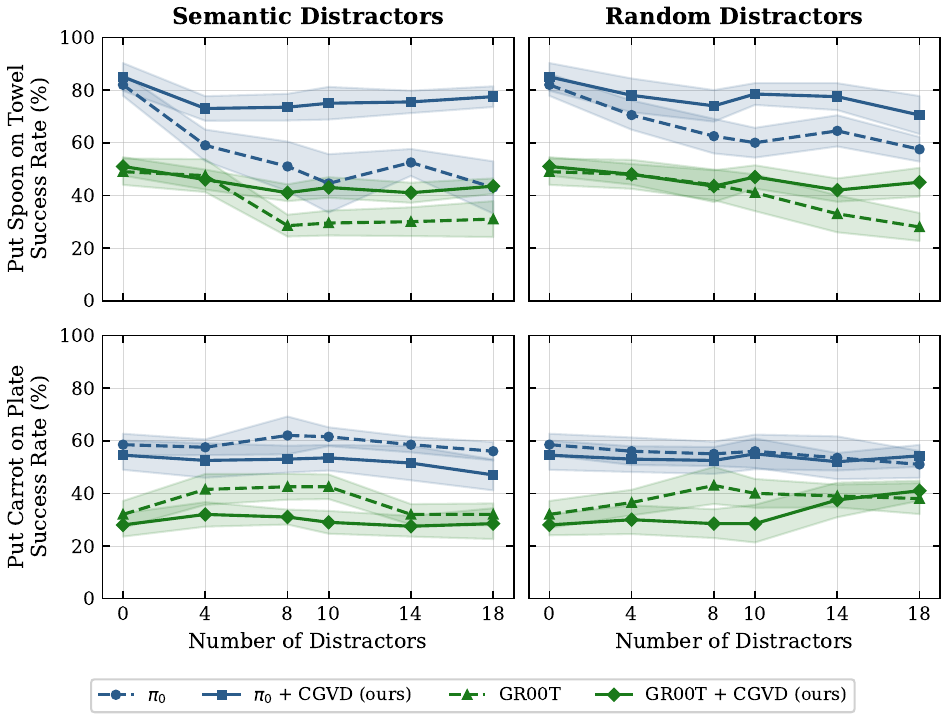}
    \caption{Success rate vs.\ number of distractors. \textbf{Left:} semantic distractors. \textbf{Right:} random distractors. \textbf{Top:} spoon on towel. \textbf{Bottom:} carrot on plate. Dashed lines represent the baseline VLA; solid lines represent +CGVD. Colors denote specific model architectures. To ensure statistical significance, each data point represents the average success rate over 200 independent evaluation rollouts (20 episodes $\times$ 10 random seeds), totaling \textbf{19,200 episodes} for the results visualized in this figure.}

  \label{fig:scaling}
\end{figure*}

\subsection{Vision-Language-Action Models}
The convergence of large language models with robotic control has yielded a new class of generalist policies known as Vision-Language-Action (VLA) models. Architectures such as RT-2 \cite{rt2} and OpenVLA \cite{openvla} leverage internet-scale vision-language pre-training to achieve remarkable zero-shot generalization to unseen objects and instructions. Similarly, policies like Octo \cite{octo} utilize transformer-based diffusion heads to aggregate behavior across diverse robot embodiments. While these models excel at general tasks with simple layouts, they often struggle with high-precision geometric grounding in cluttered scenes. This limitation stems partly from the discretization of action spaces \cite{rt2, peract} and the static patch resolution inherent in Vision Transformer (ViT) backbones \cite{vit_analysis}. Without coarse-to-fine zooming mechanisms \cite{c2f_q_attention}, these architectures are susceptible to feature dilution \cite{distracted_robot}, where high-frequency background noise competes with task-relevant signals \cite{intriguing_vit}.

\subsection{Robustness via Adaptation and Data Scaling}
Recent advances in Vision Foundation Models (VFMs) have decoupled perception from specific task training. Architectures such as SAM 3 \cite{sam3} and GroundingDINO \cite{groundingdino} enable zero-shot object localization based on free-form text queries. In robotics, these tools have been primarily employed to generate 3D value maps \cite{voxposer}, construct semantic maps \cite{conceptfusion}, or label datasets for offline training \cite{robogen}. However, these approaches largely utilize VFMs to \textit{add} information, often highlighting target regions via visual prompting \cite{RoboticVisualInstruction}. Our work inverts this paradigm: we leverage the open-set discriminative power of VFMs to identify and \textit{suppress} task-irrelevant regions. By explicitly subtracting non-causal pixels, our framework acts as a semantic information bottleneck \cite{BC-IB}, serving as a high-pass filter that blocks clutter while preserving the geometric signals essential for the downstream VLA.

\subsection{Robustness via Adaptation and Domain Randomization}
Addressing visual clutter in robotic manipulation has traditionally relied on domain randomization \cite{Tobin2017} or massive-scale pre-training on diverse datasets \cite{bridgedata, open_x_embodiment}. While modern VLAs leverage this internet-scale data for semantic generalization, they remain fragile to distributional shifts, particularly when distinguishing targets from high-frequency background noise \cite{colosseum, robocasa}. To mitigate this, adaptation methods \cite{obeyed} and fine-tuning strategies train specialized attention layers to recover geometric grounding. However, these approaches are resource-intensive, often necessitating task-specific data collection and incurring significant computational overhead. In contrast, our framework operates at inference time, extending the capabilities of frozen foundation models without parameter updates.

\subsection{Inference-Time Attention Intervention}
A growing body of work attempts to correct VLA behavior during inference. The Distracting Token Pruning (DTP) \cite{dtp} detects and suppresses visual tokens that do not align with the text prompt. DTP utilizes a 'soft' pruning strategy in feature space. However, this approach struggles when distractors share semantic features with the target, as the initial ViT self-attention layers may already be entangled \cite{intriguing_vit}. Our approach differs by intervening in pixel space. Unlike Visual Prompting \cite{RoboticVisualInstruction} which adds information to guide attention, we utilize vision foundation models to remove information, preventing attention leakage to distractors.

\section{Methodology}
\label{sec:methodology}
We present Concept-Gated Visual Distillation (CGVD), a training-free, model-agnostic inference-time framework that selectively removes distractor objects from a robot's visual observations while preserving task-relevant entities. CGVD operates as a perception wrapper around any Vision-Language-Action (VLA) policy, requiring no fine-tuning or policy modification. The key insight is that a language instruction already specifies which objects matter; CGVD leverages this signal to gate visual content, allowing only task-relevant information through to the downstream policy.

\subsection{Problem Formulation}
\label{sec:problem}

Consider a manipulation task specified by a language instruction $l$ (e.g., ``put spoon on towel'') executed by a VLA policy $\pi(a_t \mid o_t, l)$ that maps an observation $o_t \in \mathbb{R}^{H \times W \times 3}$ and instruction to an action $a_t$. In cluttered environments, $o_t$ contains distractor objects that are semantically or visually confusable with the task-relevant objects. These distractors degrade $\pi$ by introducing ambiguity in visual grounding. For instance, a spatula may be grasped instead of a spoon.

CGVD defines a distillation function $\phi$ that produces a clean observation $\hat{o}_t = \phi(o_t, l)$ in which distractors are replaced with background while the target objects and robot arm are preserved. The downstream policy then acts on the distilled observation: $a_t = \pi(\hat{o}_t, l)$. Because $\phi$ is applied at inference time and interacts with $\pi$ only through the observation interface, it is agnostic to the policy architecture.

\subsection{Concept-Gated Decomposition}
\label{sec:decomposition}

CGVD begins by parsing the language instruction $l$ to extract a target concept $c_{\text{tgt}}$ and an anchor concept $c_{\text{anc}}$. For instance, ``put spoon on towel'' yields $c_{\text{tgt}} = \texttt{spoon}$ and $c_{\text{anc}} = \texttt{towel}$. These concepts define two complementary sets that partition the scene:
\begin{itemize}
    \item The \textbf{safe set} $\mathcal{S} = \{c_{\text{tgt}},\; c_{\text{anc}},\; \text{robot}\}$: entities that must remain visible.
    \item The \textbf{distractor set} $\mathcal{D} = \{d_1, \ldots, d_K\}$: semantic categories that may appear as clutter (e.g., spatula, fork, knife).
\end{itemize}

This language-grounded decomposition is what makes the approach concept-gated: the instruction determines the gate, and only objects outside the gate are candidates for removal. Unlike prior work that relies on large language models to determine object relevance~\cite{byovla}, our parsing is deterministic and requires no additional API.

\subsection{Text-Prompted Instance Segmentation}
\label{sec:segmentation}
Both the safe set and distractor set are segmented from the observation using SAM3~\cite{sam3}. The two sets produce independent mask channels:
\begin{align}
    M_{\text{dist}} &= \textstyle\bigcup_{d_k \in \mathcal{D}} \textsc{Seg}(o_t,\, d_k), \label{eq:distractor_mask} \\
    M_{\text{safe}} &= \textsc{Seg}(o_t,\, c_{\text{tgt}}) \,\cup\, \textsc{Seg}(o_t,\, c_{\text{anc}}), \label{eq:safe_mask}
\end{align}
where $\textsc{Seg}(o_t, c)$ denotes the union of all instance masks returned by SAM3 for concept $c$ in observation $o_t$. An important computational optimization is that the vision encoder is executed only once for the initialization frame ($t=0$), with its masks reused across all frames.

\subsection{Two-Layer Target Refinement}
\label{sec:refinement}
A fundamental limitation of open-set segmentation models is that they evaluate text prompts independently. Consequently, a single-pass detection often suffers from semantic confusion, where visually similar distractors are misidentified as the target (e.g., a spatula yielding high confidence for the prompt ``spoon''). To ensure the true target survives the distillation process without relying on the VLA's flawed soft-attention, we introduce a necessary two-layer refinement pipeline on the initial frame ($t=0$).

\textbf{Layer 1: Cross-Validation.} To mathematically penalize distractors, we compute a genuineness score $g(s_i)$ for each target instance $s_i$. This measures the confidence differential between its safe-set and distractor identities:
\begin{equation}
    g(s_i) = \sigma_{\text{safe}}(s_i) \;-\; \max_{\substack{d_j \in \mathcal{D} \\ \text{IoU}(s_i, d_j) > \eta}} \sigma_{\text{dist}}(d_j),
    \label{eq:genuineness}
\end{equation}
where $\sigma_{\text{safe}}$ and $\sigma_{\text{dist}}$ are object confidences, and $\eta$ is an IoU threshold. Genuine targets yield $g > 0$, while imposters yield $g < 0$. Negative values are explicitly preserved to actively penalize false positives in the next layer.

\textbf{Layer 2: Spatial Disambiguation.} Even after cross-validation, the target mask may contain fragmented artifacts or multiple disjoint physical objects. To isolate the correct entity, we evaluate each connected component $C_k$ using a composite score:
\begin{equation}
    \text{score}(C_k) = \bigl(1 + g^*(C_k)\bigr) \cdot \sigma^*(C_k).
    \label{eq:layer3}
\end{equation}
Here, $g^*(C_k)$ is maximum genuineness, and $\sigma^*(C_k)$ is peak safe-set confidence. These factors jointly favor genuine and high-confidence components. Only the top-scoring component is retained.

To illustrate the necessity of this pipeline, consider a scene with a target \texttt{spoon} and a \texttt{spatula} distractor. If model misidentifies the spatula as a ``spoon'' ($\sigma_{\text{safe}}=0.6$) but correctly detects it as a ``spatula'' ($\sigma_{\text{dist}}=0.9$), its genuineness drops to $-0.3$. Consequently, its Layer 2 composite score is heavily penalized ($(1 - 0.3) \cdot 0.6 = 0.42$). This allows the true spoon (which maintains a high positive genuineness) to outscore the imposter and be successfully isolated.

\subsection{Concept-Gated Mask Composition}
\label{sec:gating}
The refined masks are combined into a final inpainting mask via set-theoretic gating:
\begin{equation}
    M_{\text{inp}} = \text{dilate}(M_{\text{dist}},\, r_d) \;\setminus\; \text{dilate}(M_{\text{safe}},\, r_s),
    \label{eq:gating}
\end{equation}
where $r_d$ is a distractor dilation radius, and $r_s \geq r_d$ is a safe-set dilation radius that creates a protective buffer. All masks are binarized with a threshold of $0.5$ before dilation to eliminate soft-value artifacts. 

\subsection{Clean Scene Generation via Inpainting}
\label{sec:inpainting}
We generate a single clean scene: An image of the workspace with distractors removed by applying LaMa \cite{lama}, a Fourier convolution-based inpainting model to initial frame. The inpainting mask is constructed as:
\begin{equation}
    M_{\text{lama}} = M_{\text{inp}} \;\cup\; \text{dilate}(M_{\text{robot}},\, r_e).
    \label{eq:inpaint_mask}
\end{equation}
LaMa fills the masked regions with photorealistic background texture, preserving spatial cues critical for manipulation. The clean scene is computed once per episode and cached for all subsequent frames.


\subsection{Temporally Consistent Compositing}
\label{sec:compositing}
At each timestep $t > 0$, the distilled observation is produced by smoothly blending the live camera frame $o_t$ with the cached clean scene $\hat{o}_{\text{clean}}$ using a Gaussian-blurred compositing mask $\alpha$. 
To make inpainting artifacts do not obscure the robot arm and compromise visual proprioception, we enforce a pixel-level overwrite of the robot onto the final composited image.
In simulation, we use SimplerEnv's ground-truth robot mask boundary that prevents any frame-to-frame compositing jitter. In real-robot deployments SAM3 can achieve the similar proprioceptive protection.

\begin{figure}[t]
    \centering
    \includegraphics[width=\linewidth]{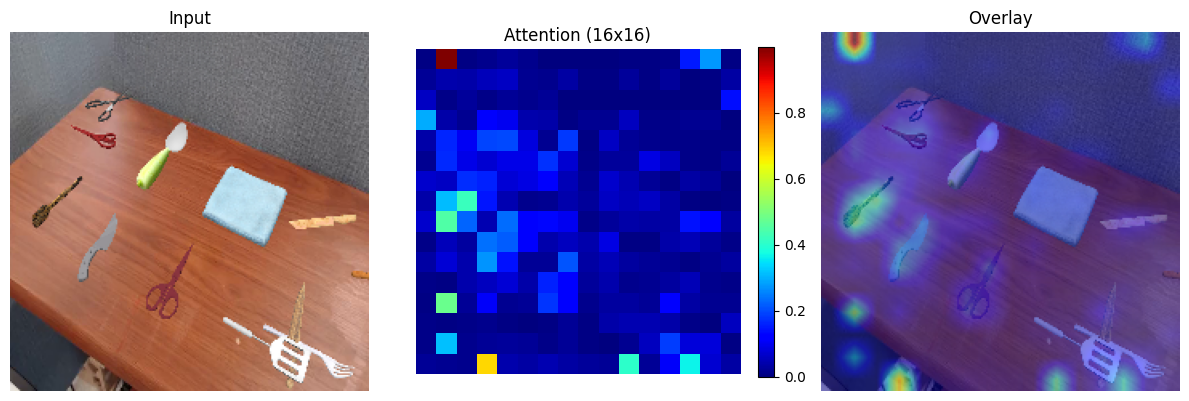}
    
    \vspace{0.1cm} 
    
    \includegraphics[width=\linewidth]{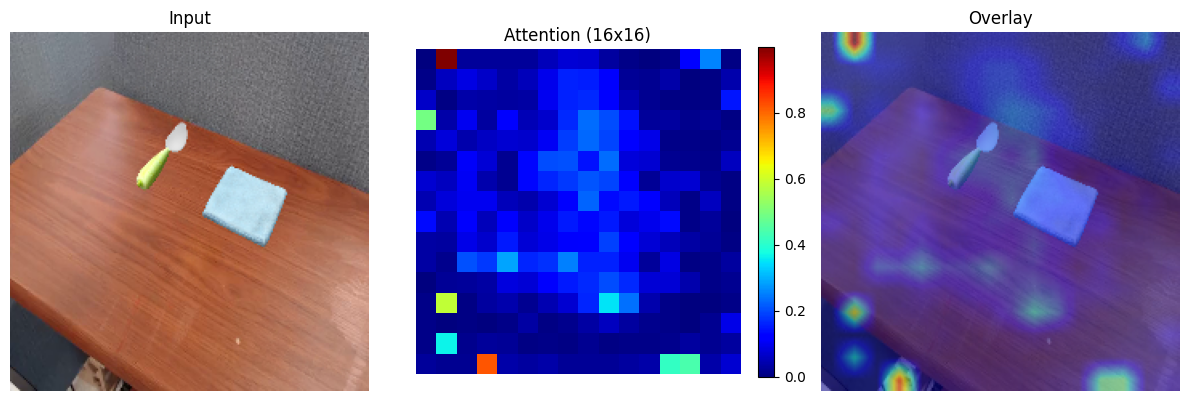}
    
    \caption{\textbf{Qualitative Analysis of Attention Repair.} 
    (Top) The baseline policy suffers from attention dispersion, focusing on the distractors rather than the spoon. 
    (Bottom) Our CGVD method inpaints the distractors, forcing the attention mechanism to collapse onto the true target.}
    \label{fig:attention_analysis}
\end{figure}

\section{Experiments}
We evaluate CGVD across two VLA architectures on tabletop manipulation tasks with controlled distractor injection. Our experiments address three questions: (1)~Does CGVD improve clutter robustness across different VLA architectures and tasks? (2)~How does performance scale with distractor density? (3)~How do individual CGVD components contribute?

\subsection{Experimental Setup}
\textbf{Environment.} We evaluate in SimplerEnv~\cite{simplerenv}, a high-fidelity simulation benchmark with demonstrated real-to-sim correlation for VLA evaluation. 
Furthermore, both core perception components of CGVD SAM3~\cite{sam3} and LaMa~\cite{lama} were independently trained and validated on real-world imagery, ensuring that the sim-to-real risk is bounded to policy-level transfer, which SimplerEnv~\cite{simplerenv} explicitly addresses.

All experiments use the WidowX robotic arm with a single fixed third-person camera, matching the Bridge dataset training setup. We test two Bridge dataset tasks: (1) spoon on towel and (2) carrot on plate

We evaluate three distractor types: (1) Semantic: objects with high semantic proximity and shared visual characteristics with the target; (2) Random: arbitrary objects with no semantic or visual similarity; and (3) Attribute: objects sharing the same category and form as the target but differing in physical properties.
Distractors are drawn from RoboCasa~\cite{robocasa} and the YCB ~\cite{ycb} dataset, and placed on the workspace using collision aware grid-based placement.

\textbf{Protocol.} We report success rate (SR) across 10 random seeds. Each condition consists of 20 episodes per seed, totaling 200 episodes per task/distractor count, with matched seeds between baseline and CGVD.



\begin{table}[!t]
\centering
\caption{\textbf{Attribute Distractor Sensitivity.} Success rates on Put spoon on towel with attribute distractors. Results are averaged over 10 random seeds (20 episodes per seed). CGVD shows stronger average robustness on both simple and complex prompts, though both methods exhibit non-monotonic variance at low distractor counts.}
\label{tab:combined_attribute_scaling}
\renewcommand{\arraystretch}{1.2} 
\resizebox{\columnwidth}{!}{%
\begin{tabular}{l ccc ccc}
\hline
& \multicolumn{3}{c}{\textbf{Simple Prompt}}  
& \multicolumn{3}{c}{\textbf{Complex Prompt}} \\
\cline{2-4} \cline{5-7}
\textbf{\# Distractors}  
  & \textbf{$\pi_0$} & \textbf{CGVD} & \textbf{$\Delta$}  
  & \textbf{$\pi_0$} & \textbf{CGVD} & \textbf{$\Delta$} \\
\hline
0 & 86.0 & \textbf{90.0} & +4.0  & 85.0 & \textbf{87.0} & +2.0 \\
1 & \textbf{80.0} & 78.0 & $-$2.0 & \textbf{74.0} & 69.0 & $-$5.0 \\
2 & 73.0 & \textbf{87.0} & +14.0 & 69.0 & \textbf{77.0} & +8.0 \\
3 & 68.0 & \textbf{75.0} & +7.0  & 64.0 & \textbf{74.0} & +10.0 \\
4 & 75.0 & \textbf{87.0} & +12.0 & 57.0 & \textbf{73.0} & +16.0 \\
\hline
\end{tabular}%
}
\end{table}

\subsection{Main Results}
Figure~\ref{fig:scaling} illustrates the task success rates as the number of distractors increases from 0 to 18. We observe distinct trends across the evaluated VLA models ($\pi_0$ and GR00T) depending on the nature of the task and clutter.

First, in the presence of semantically confusing clutter, baseline performance degrades precipitously. Here, applying CGVD successfully prevents this performance collapse, maintaining a significantly higher success rate floor. The performance gap between the baselines and CGVD widens as the environment becomes more cluttered, demonstrating CGVD's robust defense against adversarial distractors.

Conversely, in the Carrot on Plate task. The baseline policies exhibit a slight performance increase at moderate distractor densities before degrading. 
This likely arises because a moderate amount of contextual clutter better aligns with the object-rich scenes present in their large-scale pre-training distributions, providing visual anchors for reasoning.

While CGVD normalizes the visual observation to a consistent state, it consistently underperforms the baseline in this specific scenario. This suggests that when a task naturally benefits from contextual clutter, aggressively masking the scene deprives the VLA of useful environmental reasoning. Furthermore, aggressive inpainting in these context-dependent tasks can occasionally introduce generative artifacts that disrupt the spatial geometry compared to the baseline. Therefore, while CGVD provides critical protection against adversarial semantic distractors, it may trade off peak performance in scenarios where background objects serve as beneficial visual anchors.

\subsection{Fine-Grained Semantic Grounding}
The precision-reasoning gap is most acute when distractors share a semantic class with the target but differ in specific attributes. In standard VLAs, visual tokens for complex queries like 'Put spoon with green handle on towel' are often reduced to a bag-of-words, causing the policy to entangle the target with fully green objects or ignore modifiers entirely. To evaluate this, we evaluate the baseline and CGVD across 0--4 random attribute distractors using two prompt structures: Simple (Put green spoon on towel) and Complex (Put spoon with green handle on towel).

Table~\ref{tab:combined_attribute_scaling} reveals a distinct trend: while the baseline performs adequately with simple adjective-noun pairs, its performance degrades significantly on compositional queries as clutter increases (dropping from 85.0\% at 0 distractors to 57.0\% at 4 distractors). Conversely, while CGVD experiences a non-monotonic variance at 1 distractor, it demonstrates superior robustness at higher distractor densities. Because SAM 3 utilizes rich contextual cues for open-set grounding, the complex prompt enforces strict attribute adherence. CGVD successfully treats attribute-conflicting objects as background, maintaining a much more stable success rate floor (73.0\% at 4 distractors) compared to the baseline's sharp decline.

\begin{table}[t!]
\centering
\caption{ \textbf{Ablation study} on the Spoon on Towel task ($\pi_0$, 18 semantic distractors). Each row removes one component from the full CGVD pipeline. To ensure statistical significance, every configuration was evaluated over 200 episodes (20 episodes across 10 seeds).}
\label{tab:ablation}
\begin{tabular}{lc}
\hline
\textbf{Configuration} & \textbf{SR (\%)} \\
\hline
Baseline (no CGVD)        & 43.0 \\
CGVD (full pipeline)               & \textbf{77.5} \\
\quad -- Mean-color fill           & 56.5 \\
\quad -- Two-layer target refinement & 65.0 \\
\quad -- Robot mask protection     & 73.0 \\
\hline
\end{tabular}
\end{table}

\subsection{Ablation Studies}
\label{sec:ablation}
To validate CGVD's structural components, we systematically ablate 
the pipeline on the \textit{Spoon on Towel} task with 18 semantic 
distractors (Table~\ref{tab:ablation}). We use $\pi_0$ as the base 
policy, as it achieves higher overall success rates and thus provides 
a more demanding testbed for isolating individual component contributions.

Removing the Two-Layer Target Refinement reduces SR from 77.5\% to 
65.0\%. Without cross-validation, the segmentation model cannot 
distinguish true targets from visually similar distractors, causing 
genuine targets to be erroneously inpainted out of the scene.
Replacing LaMa with a mean-color fill incurs the largest single 
drop (to 56.5\%), as the stark, unnatural region boundaries act as 
adversarial patches to the VLA's ViT backbone, directly disrupting 
planning. Finally, removing Robot Mask Protection reduces SR 
to 73.0\%; without stable visual proprioception, the compositing 
mask occasionally occludes the robot arm, producing erratic 
trajectories.

\subsection{Latency Analysis}
CGVD optimizes for real-time control by executing computationally expensive operations on the initialization frame ($t=0$). For $t>0$, the system performs lightweight image compositing using the cached background. As shown in Table~\ref{tab:latency}, this strategy adds negligible overhead during execution, maintaining the VLA's native control frequency.

 \begin{table}[htbp]
  \centering
  \caption{\textbf{System Latency.} CGVD concentrates segmentation and inpainting at $t{=}0$; runtime compositing adds moderate overhead
  at the VLA control frequency.}
  \label{tab:latency}
  \begin{tabular}{lcc}
  \hline
  \textbf{Phase} & \textbf{Base $\pi_0$ (ms)} & \textbf{CGVD (ms)} \\
  \hline
  Initialization ($t=0$) & --- & $4{,}914$ \\
  Execution ($t>0$)      & $317$ & $421$ \\
  \hline
  \end{tabular}
  \end{table}

\section{Limitations}

While CGVD effectively mitigates semantic clutter, it relies on two key assumptions. First, the static background: our clean scene generation caches the inpainted background after the initialization frame. If a distractor is moved dynamically, the cached background will desynchronize from the physical scene. Although real-time mask updating could address this, continuously querying CGVD introduces latency that is currently prohibitive for high-frequency, real-time robotic control, making our cached approach a practical trade-off.

Second, inpainting fidelity in non-semantic clutter. As observed in the Carrot task, aggressive inpainting of clutter can inadvertently lead to a slight degradation in task success rates compared to the baseline.

Finally, while the inference overhead is minimized via caching, the single-frame initialization introduces a brief startup latency before the first action, though this is negligible compared to the robot's mechanical movement time.

\section{Conclusion}
In this paper, we introduced Concept-Gated Visual Distillation (CGVD), a training-free inference framework designed to bridge the Precision-Reasoning Gap in VLA models. By explicitly leveraging language-grounded segmentation and Fourier-based inpainting, CGVD isolates target objects and suppresses semantic distractors without requiring architectural modifications. Extensive evaluations demonstrate that CGVD prevents performance collapse in cluttered environments and improves generalization to out-of-distribution targets. While bounded by static background assumptions, our approach establishes visual distillation as a highly efficient prerequisite for deploying foundation models in unstructured manipulation tasks. Future work will explore real-time mask updating to handle interactive clutter.








\newpage

\bibliographystyle{IEEEtran}
\bibliography{references}

\end{document}